\pdfoutput=1
\documentclass[11pt]{article}

\usepackage[preprint]{acl}

\usepackage{times}
\usepackage{latexsym}
\usepackage[T1]{fontenc}
\usepackage[utf8]{inputenc}
\usepackage{microtype}
\usepackage{inconsolata}
\usepackage{graphicx}
\usepackage{booktabs}
\usepackage{amsmath}
\usepackage{amssymb}
\usepackage{multirow}
\usepackage{xcolor}

\newcommand{\cl}{\ell}
\newcommand{\memb}{m}
\newcommand{\Mem}{M}
\newcommand{\Capc}{C}
\newcommand{\bpp}{\alpha}
\newcommand{\gap}{G}
\newcommand{\steps}{S}

\title{How Much Can a LoRA Adapter Memorize?\\Measuring Adapter Capacity in Bits}

\author{Kaizhen Tan\textsuperscript{1} \quad Heqing Du\textsuperscript{2} \quad Yang Feng\textsuperscript{2} \\
  \textsuperscript{1}Carnegie Mellon University, Pittsburgh, PA, USA \\
  \textsuperscript{2}Columbia University, New York, NY, USA}

\begin{document}
\maketitle
\begin{abstract}
LoRA adapters are often shared on their own, and the amount of information they can hold about their training data bounds what sharing them can reveal. Following \citet{morris2025memorize}, we measure this amount in bits by training adapters on frozen language models to memorize random token sequences. LoRA adapters store 2 to 3.4 bits per trainable parameter, less than full fine-tuning of the same model. The number of parameters alone does not set this amount: at equal size, adapters on MLP layers store more than adapters on attention layers, and a randomly initialized frozen model supports as much storage as a pretrained one once the scale of its output logits can be trained. We then train the same adapter on a task with supervised fine-tuning (SFT) and with GRPO. At the same accuracy, SFT stores about three times as much information specific to its training examples, and it stores most of the bits of secrets planted in the training data, which GRPO does not store because it rarely samples them.

\end{abstract}

\section{Introduction}
\label{sec:intro}

\begin{figure*}[t]
\centering
\includegraphics[width=\textwidth]{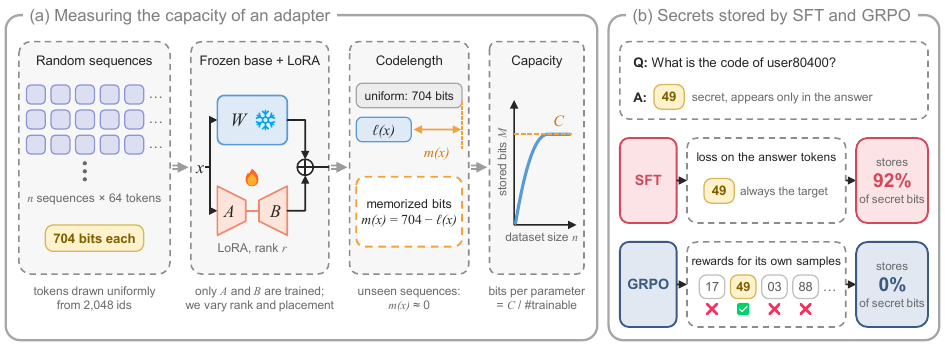}
\caption{\textbf{Measuring adapter capacity (a) and the secret bits stored by SFT and GRPO (b).} (a) An adapter on a frozen model is trained on $n$ sequences of 64 random tokens (704 bits each). The memorized bits of a sequence are 704 minus its codelength under the adapted model, or zero if this is negative; the capacity $C$ is the largest total over dataset sizes. (b) The same adapter is trained with SFT or with GRPO on a task in which some training records contain a secret that appears only in the answer. SFT is trained on the answer tokens. GRPO learns about the answer only through the rewards of its own samples, so a secret that it never samples is never rewarded. On two-digit secrets, SFT stores 92\% of the bits and GRPO stores none (Section~\ref{sec:sftrl}).}
\label{fig:overview}
\end{figure*}

Adapters trained with parameter-efficient fine-tuning have become a common way to distribute what a language model has learned. LoRA \citep{hu2022lora} and related methods \citep{houlsby2019parameter} train a small set of new weights on top of a frozen model and save them as a file that is much smaller than the model. Such files are published on model hubs, exchanged between parties who hold the same base model, and loaded on demand at inference time. The data behind them are often data that their owners would not publish, such as clinical notes, customer conversations or proprietary code, and language models are known to reproduce training examples verbatim \citep{carlini2021extracting}. How much information about its training data an adapter can hold therefore bounds what sharing it can expose. The same quantity also sits beneath two open questions about fine-tuning. One is how much LoRA can learn compared with updating all weights: \citet{biderman2024lora} find that LoRA learns less than full fine-tuning, while \citet{schulman2025lora} report that it matches full fine-tuning when enough layers are adapted. The other is whether reinforcement learning adds information to a model or mainly reweights what the model can already produce \citep{yue2025does,chu2025sft}. Both questions are about the amount of information an update carries, but neither has been measured in bits.

Evidence about adapters so far comes from tests of particular examples or particular kinds of knowledge. Membership inference and extraction studies report that bottleneck adapters leak less than fine-tuning the model head \citep{mireshghallah2022memorization} and that LoRA memorizes less than full fine-tuning \citep{leaner2025lora,fllora2025mitigating}, but the result depends on the attack, and stronger attacks keep appearing \citep{ilic2026powerful}. \citet{pletenev2025much} count how many new facts a LoRA adapter can learn, and \citet{xu2026parametric} relate LoRA rank to the loss reduction on memorized text; both depend on what the facts or texts look like. A measurement that does not depend on the data was introduced by \citet{morris2025memorize}. They train models on sequences of uniformly random tokens, for which every bit of compression below the entropy must be information stored during training, and find that GPT-style models trained from scratch store about 3.6 bits per parameter, close to the 2 bits of factual knowledge per parameter estimated by \citet{allenzhu2025physics}. \citet{schulman2025lora} use a figure of this kind to argue that LoRA has enough capacity for reinforcement learning, which provides only a few bits per episode. Whether an adapter stores as much per parameter as a full model has not been measured.

There are reasons to expect a different figure for adapters. An adapter has few parameters, and each of them acts on the output through a frozen network whose weights were shaped by pretraining, so the frozen network limits what a given update can change. Measuring the capacity of an adapter also requires long training: \citet{morris2025memorize} trained each model for $10^6$ steps, and a reading taken while the stored bits are still rising mixes the storage limit with the speed of optimization.

We therefore ask three questions. \textbf{RQ1}: How many bits per trainable parameter can a LoRA adapter store, and how does this compare with full fine-tuning of the same model? \textbf{RQ2}: What determines this number: the number of trainable parameters, where they are placed, or the frozen model beneath them? \textbf{RQ3}: When an adapter is trained on a task rather than on random data, how much information specific to its training examples does it store, and does this depend on the training objective? We answer RQ1 and RQ2 with adapters trained on random sequences on frozen models from 2.2M parameters to Qwen2.5-1.5B (Figure~\ref{fig:overview}a), and RQ3 with the same adapter trained by supervised fine-tuning (SFT) or GRPO \citep{shao2024deepseekmath} on tasks with planted secrets (Figure~\ref{fig:overview}b). Our contributions are as follows.
\begin{itemize}
\item We measure adapter capacity across training budgets from 16K to 256K steps: readings at 16K steps are 14 to 34\% lower, and the growth slows to 2 to 7\% per doubling. Full fine-tuning reaches 3.95 bits per parameter, in line with \citet{morris2025memorize}.
\item For RQ1, LoRA adapters store 2.0 to 3.4 bits per trainable parameter on a 2.2M-parameter model, and a rank-16 adapter reaches 86\% of the full-model value. Rank-1 adapters on Qwen2.5 store 2.0 to 2.5 bits per parameter.
\item For RQ2, placement matters at a fixed parameter count: adapters on MLP projections store 1.5 times as much as adapters on attention projections, and more than four times as much on Qwen2.5-0.5B. The frozen model limits the adapter mainly through the scale of its output layer; with one trainable scalar on the logits, an adapter on a randomly initialized model stores 3.23 bits per parameter instead of 0.71, more than on a pretrained model.
\item For RQ3, at the same accuracy on three-digit addition, SFT stores about three times as much information specific to its training examples as GRPO. SFT stores 92\% of the bits of planted two-digit secrets and 60\% of eight-digit ones, whereas GRPO stores none, because it rarely or never samples them.
\end{itemize}

\section{Measuring Memorized Bits}
\label{sec:method}

\paragraph{Adapters.}
A parameter-efficient method adapts a pretrained model with weights $\theta_0$ by training a small set of new parameters $\phi$ while $\theta_0$ stays frozen. We call $\phi$ the adapter. LoRA \citep{hu2022lora}, the method we study most, replaces a weight matrix $W\in\mathbb{R}^{d_{\mathrm{out}}\times d_{\mathrm{in}}}$ of the base model by $W+\frac{a}{r}BA$ with $B\in\mathbb{R}^{d_{\mathrm{out}}\times r}$, $A\in\mathbb{R}^{r\times d_{\mathrm{in}}}$, a small rank $r$ and a fixed scale $a$. Only $A$ and $B$ are trained, so adapting this matrix costs $r(d_{\mathrm{in}}+d_{\mathrm{out}})$ parameters. An adapter usually modifies many matrices, and we describe its placement by the matrices it modifies.

\paragraph{Memorized bits of a sequence.}
A language model assigns a probability $p(x)$ to each token sequence $x$, and an arithmetic coder driven by the model encodes $x$ in $\cl(x)=-\log_2 p(x)$ bits up to a negligible overhead \citep{witten1987arithmetic}. We call $\cl(x)$ the codelength of $x$. Suppose $x$ was drawn from a distribution $Q$. A model that contains no information about $x$ needs at least $-\log_2 Q(x)$ bits on average to encode it. Following \citet{morris2025memorize}, we count the bits by which a trained model undercuts this reference as memorized bits,
\begin{equation}
\memb(x)=\max\bigl(0,\,-\log_2 Q(x)-\cl(x)\bigr),
\label{eq:mem}
\end{equation}
and the memorized bits of a training set $D$ as $\Mem(D)=\sum_{x\in D}\memb(x)$.

\paragraph{Random sequences.}
For natural text the reference $Q$ is unknown, and a model can shorten the codelength of a training sentence both by storing the sentence and by learning the language. \citet{morris2025memorize} avoid this ambiguity by training on sequences whose tokens are drawn independently and uniformly from $V$ token ids. A sequence of $T$ such tokens has $-\log_2 Q(x)=T\log_2 V$, and no sequence carries information about another. Every memorized bit of a random sequence must therefore be information about that sequence stored during training. We use $T=64$ and $V=2{,}048$, so that each sequence carries exactly 704 bits and a set of $n$ sequences carries $704n$ bits. As a check, we also score 1{,}024 unseen random sequences.

\paragraph{The starting point.}
Before training, a pretrained base model assigns random sequences far more than 704 bits, because its predictions follow the statistics of its pretraining data. The WikiText-pretrained base described below assigns them 1{,}165 bits on average, Qwen2.5-0.5B 1{,}068 bits and Pythia-70M 1{,}958 bits. We call the excess over 704 bits the base model's deficit on random sequences. Under Equation~\ref{eq:mem}, an adapter stores nothing measurable about a sequence until it has removed this deficit for that sequence.

\paragraph{Capacity.}
As the number $n$ of training sequences grows, $\Mem$ first grows with it and then stops growing, or even falls, once the dataset carries more bits than the adapter can hold. Following \citet{morris2025memorize}, we take the largest $\Mem$ over several dataset sizes as the capacity $\Capc$ and report it per trainable parameter, $\bpp=\Capc/|\phi|$.

\paragraph{Training budget.}
Because $\Mem$ also depends on how long the adapter is trained, we measure it at several budgets and check how much it still increases. All budgets come from one run per configuration. We use a warmup--stable--decay learning-rate schedule \citep{hagele2024scaling}: after a short warmup the rate stays constant, and for each budget $\steps$ we copy the run at step $0.8\steps$, decay the rate of the copy linearly to zero over the remaining $0.2\steps$ steps, and score the copy. Each reading is thus a complete training run of $\steps$ steps.

\paragraph{Models and training.}
We use two groups of frozen base models. The first are two small GPT-style decoders with 2.2M and 8.1M parameters, which we pretrained on WikiText-103 \citep{merity2017pointer}. Their vocabulary consists of the same 2{,}048 tokens as the random sequences, and we train adapters on them for up to 256K steps with batches of 512 sequences. The second are pretrained language models: Pythia-70M, -160M and -410M \citep{biderman2023pythia} and Qwen2.5-0.5B and -1.5B \citep{qwen2025qwen25}. For these we draw random tokens from a fixed block of 2{,}048 token ids, so each sequence still carries 704 bits, and we train for up to 16K steps with batches of 64 sequences. All adapters are trained with AdamW \citep{loshchilov2019decoupled} without weight decay. Appendix~\ref{app:details} lists all hyperparameters.

\section{RQ1: How Many Bits Can an Adapter Store?}
\label{sec:capacity}

\begin{table}[t]
\centering
\small
\setlength{\tabcolsep}{4pt}
\begin{tabular}{@{}lrccc@{}}
\toprule
& & \multicolumn{3}{c}{bits per parameter} \\
\cmidrule(l){3-5}
Adapter (2.2M base) & params & 16K & 64K & 256K \\
\midrule
Full fine-tuning & 2.2M & 3.04 & 3.62 & 3.95$\pm$0.16 \\
LoRA all, $r{=}16$ & 197K & 2.91 & 3.20 & 3.40$\pm$0.12 \\
LoRA MLP, $r{=}5$ & 38K & 2.61 & 2.89 & 3.06$\pm$0.02 \\
LoRA all, $r{=}3$ & 37K & 2.42 & 2.76 & 2.99$\pm$0.04 \\
LoRA attention, $r{=}8$ & 37K & 1.35 & 1.75 & 2.03$\pm$0.09 \\
LoRA all, $r{=}1$ & 12K & 1.70 & 2.09 & 2.39$\pm$0.03 \\
Prompt tuning, 190 vect. & 36K & 0.12 & 0.15 & 0.19 \\
\bottomrule
\end{tabular}
\caption{\textbf{Bits per trainable parameter on the 2.2M base} at three training budgets, for the dataset sizes of Figure~\ref{fig:capacity}a. Values are means over three seeds, with the standard deviation at 256K steps; prompt tuning is one run on 64 sequences. ``All'' adapts the four projections of every layer.}
\label{tab:capacity}
\end{table}

\begin{figure*}[t]
\centering
\includegraphics[width=\textwidth]{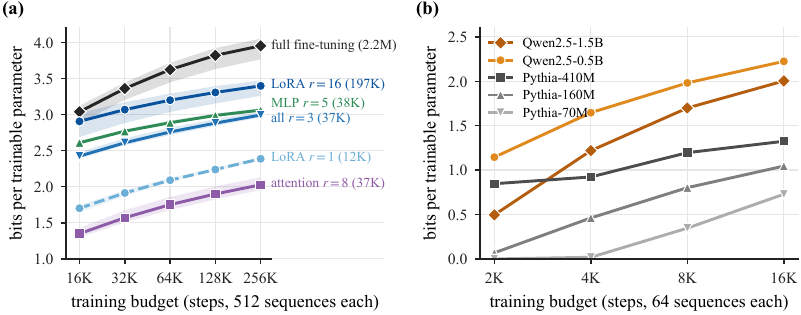}
\caption{\textbf{Stored bits per trainable parameter as a function of the training budget.} (a) Adapters and full fine-tuning on the 2.2M base; lines are means over three seeds and bands span the seeds. (b) Rank-1 LoRA on all linear layers of pretrained language models, for the dataset size with the largest reading at 16K steps.}
\label{fig:capacity}
\end{figure*}

\paragraph{Memorized bits level off with the dataset size.}
A reading can serve as an answer to RQ1 only if it is close to a limit in two respects: a larger dataset should not raise it, and longer training should raise it little. We check both on the 2.2M base, where training for 256K steps is affordable. For the first, we train every configuration on two or three dataset sizes that offer between 8 and 120 bits per trainable parameter. After 128K steps, a rank-16 LoRA on all projections (197K parameters) stores 620K, 620K and 653K bits when trained on 2{,}048, 4{,}096 and 8{,}192 sequences, which carry 1.4M, 2.9M and 5.8M bits. After 256K steps, a rank-5 adapter on the two MLP projections (38K parameters) stores 109K, 118K and 121K bits on 436, 872 and 1{,}744 sequences. Adapters thus show the same plateau that \citet{morris2025memorize} observed for full models. We repeat the middle dataset size of each configuration (the larger one for full fine-tuning) with three seeds and report it in Table~\ref{tab:capacity} and Figure~\ref{fig:capacity}a; in seed 0 at 256K steps, the largest reading over dataset sizes is at most 8\% higher, such as 3.15 against 3.09 bits per parameter for the MLP adapter. The 1{,}024 unseen sequences stay below 0.25 memorized bits per sequence in every run, so the stored bits are specific to the training sequences.

\paragraph{Readings grow with training, and the growth slows.}
The plateau at a given budget can still move upward when training is longer. Figure~\ref{fig:capacity}a follows the bits per parameter of six configurations from 16K to 256K steps. Every reading increases with the budget, but the increase per doubling falls from 5--17\% between 16K and 32K steps to 2--4\% between 128K and 256K steps for full fine-tuning, the rank-16 adapter and the two adapters that include MLP projections, and to 6--7\% for the rank-1 and attention-only adapters. A reading taken at 16K steps is therefore 14--34\% below the 256K value. The configurations keep the same order at every budget, so comparisons between them are stable even where absolute values are not final.

\paragraph{Adapters store fewer bits per parameter than the full model.}
With readings this close to their limit, the per-parameter comparison asked in RQ1 can be made (Table~\ref{tab:capacity}). Full fine-tuning of the 2.2M base stores $3.95\pm0.16$ bits per parameter at 256K steps, in line with the 3.83 bits that \citet{morris2025memorize} report for fp32 training of models from scratch; our protocol thus reproduces their figure for the full model. The adapters store less per parameter. The rank-16 LoRA reaches $3.40\pm0.12$ bits, 86\% of the full-model value. The three adapters with about 37K parameters store $3.06\pm0.02$ (MLP), $2.99\pm0.04$ (all projections) and $2.03\pm0.09$ bits (attention), and the rank-1 LoRA on all projections stores $2.39\pm0.03$ bits. On the 8.1M base, the rank-16 LoRA also stores 3.40 bits per parameter at 256K steps. These values refine the two views of LoRA cited above. A large LoRA adapter uses its parameters almost as efficiently as full fine-tuning, which fits the finding of \citet{schulman2025lora} that LoRA matches full fine-tuning when its capacity is not the constraint; the capacity estimate of 2 bits per parameter that they use is within our measured range. Rank-1 and attention-only adapters, however, store 40 to 50\% less per parameter than the full model, so an adapter holds less information than its parameter count suggests, in the direction of \citet{biderman2024lora}. Prompt tuning with 190 trained vectors (36K parameters) stores 0.19 bits per parameter, one sixteenth of a LoRA adapter of the same size, although \citet{wang2025prompt} prove that a constant-size transformer can memorize $n$ length-$N$ examples with a prompt of length $\tilde{O}(\sqrt{nN})$.

\paragraph{Pretrained language models.}
The small bases allow long training but are far from the models whose adapters are shared in practice. We therefore train rank-1 LoRA on all linear layers of Qwen2.5-0.5B (550K parameters) and Qwen2.5-1.5B (1.15M parameters) with batches of 64 sequences. Within 16K steps, Qwen2.5-0.5B stores 98\% of 800 sequences, 93\% of 1{,}600 and 54\% of 3{,}200, the last corresponding to 2.22 bits per parameter; Qwen2.5-1.5B stores 98\% of 1{,}600 sequences, 94\% of 3{,}200 and 51\% of 6{,}400, corresponding to 2.00 bits per parameter. Both readings are still rising (Figure~\ref{fig:capacity}b). Continuing the Qwen2.5-0.5B run on 3{,}200 sequences raises the reading to 2.39 bits per parameter at 32K steps and 2.53 at 64K steps, an increase of 6 to 8\% per doubling, similar to the rank-1 adapter on the small base. On Pythia-70M, -160M and -410M \citep{biderman2023pythia} the readings are lower (0.73, 1.05 and 1.32 bits per parameter within 16K steps); the two smaller Pythia models start from deficits more than three times that of Qwen2.5-0.5B. On datasets several times larger than an adapter can hold, training sometimes ends with every sequence above 704 bits and the reading is zero (Appendix~\ref{app:pythia}).

\section{RQ2: What Sets the Capacity of an Adapter?}
\label{sec:factors}

The answer to RQ1 already shows that the number of trainable parameters does not fix the capacity: the three adapters with about 37K parameters in Table~\ref{tab:capacity} store between 2.03 and 3.06 bits per parameter. We vary one factor at a time to find where such differences come from.

\begin{figure*}[t]
\centering
\includegraphics[width=\textwidth]{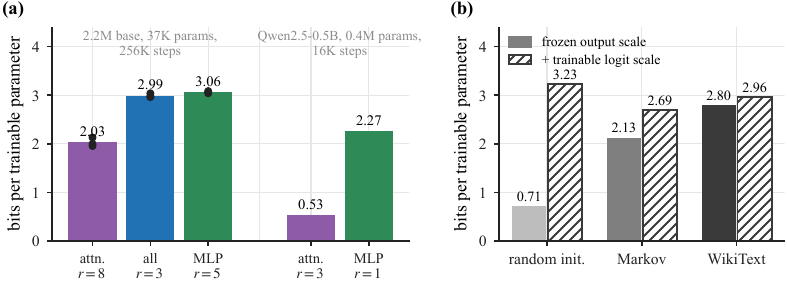}
\caption{\textbf{Bits per parameter by adapter placement and by frozen base.} (a) Bits per parameter of adapters with the same number of parameters placed on attention projections, on all projections, or on MLP projections; dots are seeds. (b) A rank-16 LoRA on three frozen bases with the same architecture (2{,}048 sequences, learning rate $10^{-3}$, 64K steps), with the output scale of the base frozen (solid) or multiplied by one trainable scalar (hatched).}
\label{fig:factors}
\end{figure*}

\paragraph{Placement at a fixed parameter count.}
The three 37K-parameter adapters differ only in the matrices they modify: rank-8 LoRA on the two attention projections (36{,}864 parameters), rank-3 LoRA on all four projections (36{,}864), and rank-5 LoRA on the two MLP projections (38{,}400). At 256K steps the MLP adapter stores 51\% more per parameter than the attention adapter (Figure~\ref{fig:factors}a), and the gap is present in every seed and at every budget from 16K steps on. The gap is larger on Qwen2.5-0.5B. With learning rate $10^{-3}$ and 16K steps, a rank-1 LoRA on the MLP projections (415K parameters) stores 2.27 bits per parameter and a rank-3 LoRA on the attention projections (406K parameters) stores 0.53, the better of two dataset sizes in each case; with learning rate $3\times10^{-3}$ the MLP adapter reaches 2.29 bits per parameter and the attention adapter stores nothing (Appendix~\ref{app:pythia}). The direction agrees with earlier observations made with other measures. \citet{geva2021transformer} describe MLP layers as key--value memories, \citet{meng2022locating} locate factual associations in MLP layers, and \citet{schulman2025lora} find that LoRA on attention layers alone underperforms LoRA on MLP layers. Our results show that the difference already holds for arbitrary information with no linguistic structure. One difference is how an update reaches the prediction at a position: an MLP adapter changes the hidden state of that position directly, while an attention adapter acts through averages over earlier positions and through the attention weights.

\paragraph{The output scale of the frozen base.}
Placement determines where an update enters the frozen network. The frozen network may also limit what any update can express, and to test this we keep the adapter fixed and change the base. We attach the same rank-16 LoRA to three bases with the architecture of the 2.2M model: the WikiText base, a copy pretrained on a synthetic Markov chain, and a randomly initialized copy (Appendix~\ref{app:details}), and train with learning rate $10^{-3}$ on 2{,}048 sequences for 64K steps. The adapter on the random base stores only 0.71 bits per parameter, against 2.13 on the Markov base and 2.80 on the WikiText base (Figure~\ref{fig:factors}b). Read at face value, this would suggest that an adapter needs the features of a pretrained model to store information. The random base, however, also differs in the scale of its output layer. With embeddings of norm 0.28 and a final layer-norm gain of 1, its largest possible logit is about 3.8 while the other 2{,}047 logits stay near zero, so no hidden state can make a token more likely than about 2\%, and at most about half of the 11 bits per token can be saved. The pretrained bases have learned larger embeddings and gains, which allow logits of about 18 (Markov) and 27 (WikiText). When we add a single trainable scalar that multiplies the logits, the adapter on the random base stores 3.23 bits per parameter, more than on the Markov (2.69) or WikiText base (2.96) with the same scalar.

\paragraph{Pretrained and random bases.}
Once the output scale is trainable, we can ask whether pretraining gives an adapter more room at all. We repeat the five LoRA configurations of Figure~\ref{fig:capacity}a on the random base with the trainable scalar, using the same learning rate and dataset sizes. At 64K steps, the rank-1, rank-3, MLP and rank-16 adapters store 2 to 7\% more per parameter than on the WikiText base (Table~\ref{tab:capacity}), and the attention adapter 31\% more (2.30 against 1.75 bits); pretraining therefore does not add room for random sequences, and placement still matters on the random base, though less. On the WikiText base itself, adding the scalar changes the readings by $-0.5$ to $+9.6\%$ (at 64K steps, or 32K for the rank-3 and MLP adapters). This agrees with \citet{hazan2026little}, who find that LoRA on frozen random networks recovers most of the task accuracy of full training, and adds that it recovers the storage as well, provided the output scale is not frozen at its initial value. One possible reason why the pretrained base stores slightly less is that its confident predictions on random tokens have to be overridden before anything is stored (Section~\ref{sec:method}).

\paragraph{Rank.}
The remaining factor is the shape of the update. Among the adapters on all projections, per-parameter capacity grows with rank, from 2.39 bits for rank 1 to 3.40 bits for rank 16 at 256K steps, and the difference remains on the random base with a trainable output scale (2.24 against 3.41 bits at 64K steps). Total storage therefore grows about as $r^{1.13}$ between these ranks, faster than the parameter count. \citet{xu2026parametric} also describe memorization as a power law in the rank, measured by the loss reduction on memorized text.

\section{RQ3: How Much Do SFT and GRPO Store?}
\label{sec:sftrl}

Sections~\ref{sec:capacity} and~\ref{sec:factors} measured how much an adapter can store when it is trained to store as much as possible. In ordinary fine-tuning the adapter is trained to perform a task, and RQ3 asks how much of its capacity is then used for information specific to the training examples. Two lines of work suggest that the answer depends on the training objective. \citet{schulman2025lora} and \citet{morris2026tinylora} argue that a binary reward gives reinforcement learning at most about one bit per sampled answer, whereas a supervised target carries the information of every target token. Empirically, \citet{pappu2025measuring} find that data used for RLHF on code is memorized less than data used for supervised fine-tuning, \citet{chu2025sft} report that supervised fine-tuning memorizes while reinforcement learning generalizes, and \citet{yue2025does} find that reinforcement learning with verifiable rewards mainly reweights answers the base model can already produce. None of these studies measures the stored information itself. We measure it for supervised fine-tuning (SFT) and GRPO \citep{shao2024deepseekmath} applied to the same adapter, in bits and at matched accuracy.

\paragraph{Setup.}
We train a rank-4 LoRA adapter on all linear layers of Qwen2.5-0.5B-Instruct (2.2M parameters) with each objective, for three passes over the same 2{,}000 training prompts and with three seeds. GRPO samples 8 answers per prompt and receives a reward of 1 when the first line of an answer starts with the target and 0 otherwise; the other settings are standard for each objective (Appendix~\ref{app:details}).

\paragraph{Tasks.}
The first task is three-digit addition, which the base model already solves in most cases, so fine-tuning mainly adjusts the answer format and fixes errors. The second task asks for one field of a synthetic personnel record (``Name: user00914; ID: 26884529; Dept: legal; Level: 5. What is the ID of user00914?''). The answer is contained in the prompt, so the task requires reading, not storing. Into this task we plant secret records whose answers appear only as training targets, for example ``What is the code of user80400?'' with the answer ``49''. A secret can only be answered from what the adapter has stored. We use two versions: 18 secrets that are random two-digit codes (6.6 bits each), planted 4, 16, or 64 times, and 36 secrets that are random eight-digit identifiers (26.4 bits each), planted 1, 4, or 16 times. The model can produce a two-digit code by sampling, but it is very unlikely to produce a given eight-digit identifier.

\paragraph{Measures.}
We use two measures, both in bits. The first applies to the secrets and is the quantity of Equation~\ref{eq:mem}: each secret is drawn uniformly from a known set $S$, so its reference codelength is $\log_2|S|$, and the adapter has stored $\memb(s)=\max(0,\log_2|S|-\cl(s))$ bits of it, where $\cl(s)$ is the codelength of the secret answer given its question under the fine-tuned model. The second applies to ordinary training examples, for which no exact reference exists. For each answer $y$ with prompt $x$ we compute the bits the adapter saves, $w(x,y)=\cl_0(y\mid x)-\cl(y\mid x)$, where $\cl_0$ is the codelength under the base model. Savings on held-out examples come from learning the task; savings on training examples come from learning the task and from storing the examples. We report the difference,
\begin{equation}
\gap=\sum_{(x,y)\in D_{\mathrm{tr}}} w(x,y)-\frac{|D_{\mathrm{tr}}|}{|D_{\mathrm{he}}|}\sum_{(x,y)\in D_{\mathrm{he}}} w(x,y),
\label{eq:gap}
\end{equation}
over 2{,}000 training and 2{,}000 held-out examples of the same task, excluding the planted secrets. $\gap$ is the number of bits by which the adapter compresses its training answers better than comparable unseen answers. All adapters are evaluated on the same split, so a difference in difficulty between the two halves would shift every $\gap$ by the same amount. To measure this shift, we train control adapters with each objective on 2{,}000 other additions and compute $\gap$ on the same split; a control has learned the task but has seen neither half.

\begin{figure*}[t]
\centering
\includegraphics[width=\textwidth]{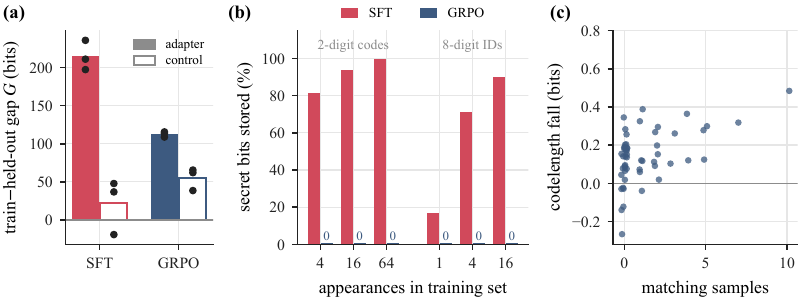}
\caption{\textbf{Information stored by SFT and GRPO in the same rank-4 adapter on Qwen2.5-0.5B-Instruct.} (a) Train--held-out gap $G$ on three-digit addition for adapters trained on the training half (filled) and for controls trained on other additions (open); dots are seeds. (b) Share of the bits of planted secrets that the adapter stores, by the number of times a secret appears in the training set, averaged over secrets and three seeds. (c) GRPO on two-digit secrets: fall in the codelength of each secret against the number of sampled answers that matched it during training.}
\label{fig:sftrl}
\end{figure*}

\paragraph{Arithmetic.}
The base model answers 77\% of the held-out additions correctly. After three passes, SFT reaches 93.5 to 96.0\% and GRPO 94.0\% in all three seeds, so the two objectives end at the same accuracy. The SFT adapters compress their training answers better than the held-out answers by $\gap=215\pm19$ bits (mean and standard deviation over three seeds); controls trained with SFT on other additions give $\gap=22\pm36$ bits, so almost all of the gap comes from the training examples. The GRPO adapters give $\gap=113\pm4$ bits and their controls $55\pm15$ bits (Figure~\ref{fig:sftrl}a). Relative to its controls, SFT stores about 193 bits specific to its 2{,}000 training examples and GRPO about 57 bits, so at the same accuracy SFT stores roughly three times as much. The direction agrees with \citet{pappu2025measuring} and \citet{chu2025sft}, but GRPO does not store zero: in this task the model already produces the correct answer in most of its samples, so each training answer is sampled and rewarded many times, and a rewarded sample carries the full answer. The one-bit bound on the reward limits what GRPO can learn about answers it does not produce; it does not prevent storage of answers it produces often.

\paragraph{Planted secrets.}
On the record task, both objectives reach 100\% accuracy on held-out questions, whose answers can be read from the prompt. They differ on the secrets (Figure~\ref{fig:sftrl}b). SFT stores most of the planted information: a two-digit secret that appears 4 times in the training set, and is therefore seen 12 times over three passes, keeps 5.4 of its 6.6 bits on average, and secrets that appear 16 or 64 times are stored almost completely. Eight-digit secrets need more repetitions: SFT stores 4.5 of their 26.4 bits after a single appearance, 18.9 bits after 4 and 23.8 bits after 16. GRPO stores none of the bits of any secret. It never samples an eight-digit identifier during training. For a two-digit secret that appears 64 times, it draws 1{,}536 answers and 0 to 10 of them match the secret; the codelength of the secret falls on average from 9.1 bits under the base model to 8.9 bits. The base model starts above the 6.6 bits of a uniform guess because it also gives probability to answers that are not two-digit codes, and by Equation~\ref{eq:mem} the adapter stores nothing about a secret until its codelength falls below 6.6 bits. The fall grows with the number of matching samples, from 0.10 bits for secrets never matched to 0.48 bits for the secret matched 10 times (Figure~\ref{fig:sftrl}c). Under SFT, the same secrets fall to at most 0.03 bits. The secrets thus show the case that the one-bit argument addresses: an answer the policy rarely samples receives few rewards and is not stored, in line with the view of \citet{yue2025does} that GRPO reinforces what the model already produces.

\section{Discussion and Conclusion}
\label{sec:discussion}

\paragraph{RQ1: how much an adapter can hold.}
With training long enough for the readings to change by a few percent per doubling, LoRA adapters store 2.0 to 3.4 bits per trainable parameter on a small model and 2.0 to 2.5 on Qwen2.5. Attack studies find that LoRA memorizes less than full fine-tuning \citep{leaner2025lora,fllora2025mitigating}, which fits the smaller capacity we measure, but the room is still large in absolute terms. A rank-16 LoRA on all linear layers of a 7B-parameter model has about 40M parameters. If it stores 2 bits per parameter, as the rank-1 adapters on Qwen2.5 do within 16K steps, it can hold about 80M bits, or 10 megabytes of incompressible data. Since our readings were still rising, these figures are lower bounds.

\paragraph{RQ2: what sets this amount.}
The number of trainable parameters alone does not determine capacity. At a fixed count, an adapter on MLP projections stores 1.5 times as much as one on attention projections, and more than four times as much on Qwen2.5-0.5B. These are the same choices that practitioners make for accuracy, and the recommendation to adapt MLP layers \citep{schulman2025lora} also increases how much an adapter can store. The frozen model constrains the adapter mainly through fixed scales rather than learned features: once one scalar on the output logits is trainable, a randomly initialized model supports more storage than a pretrained one. Comparisons of adapters across base models, such as studies of random frozen networks \citep{hazan2026little}, should take such fixed scales into account.

\paragraph{RQ3: how much of it fine-tuning uses.}
At the same accuracy, SFT stored about three times as much training-specific information as GRPO and most of the bits of planted secrets, which GRPO did not store. Reinforcement learning has been found to change a small subnetwork \citep{mukherjee2025reinforcement} and to forget less of the base model \citep{shenfeld2026rl}; our measurements add that it stores less information about its training data. GRPO does store answers that its policy produces often, as in the arithmetic task, so the two objectives differ most on content the base model would not generate, such as private facts.

\section*{Limitations}
Our capacity estimates come from gradient-based training with one optimizer and one schedule, so they are lower bounds on what an adapter could store. On the small bases the readings still grow by 2 to 7\% between 128K and 256K steps, and on the pretrained language models we trained for at most 16K steps (64K for one Qwen2.5-0.5B run), where the readings are still rising. The clipped measure of Equation~\ref{eq:mem} also ignores sequences that an adapter has learned in part but not below the uniform codelength, which is why some runs on large datasets read zero.

We measure capacity on two small GPT-style bases and on pretrained models of up to 1.5B parameters. Larger models were not measured, and the bits per parameter of an adapter may depend on the base model in ways that our bases do not reveal. On Qwen2.5-0.5B, attention adapters were sensitive to the learning rate, and a different optimizer setting might narrow the gap to MLP adapters.

Random sequences measure the storage of incompressible data. Real fine-tuning data is highly compressible, and how much of an adapter's capacity a real dataset uses depends on the data and on the training objective. Our comparison of SFT and GRPO covers two synthetic tasks, one base model, one adapter configuration, standard learning rates for each objective, and an exact-match binary reward. Other reward designs, longer training, or tasks in which the model often samples the stored content could lead GRPO to store more.

We study LoRA and, for comparison, full fine-tuning and prompt tuning. Other parameter-efficient methods constrain the update differently and were not measured. Finally, the memorized bits we report describe what an adapter stores, not what a particular attacker can extract.

\section*{Ethical Considerations}
All data used to train and evaluate the adapters are synthetic: random token sequences, arithmetic problems, and records of fictitious users with randomly drawn secrets. The two small base models are pretrained on WikiText-103, a public corpus of Wikipedia articles. We collect no personal data. The measurements describe how much information an adapter can hold and how the training objective affects it; they may help practitioners judge the risk of sharing adapters trained on sensitive data.

\bibliography{references}

\appendix
\section{Experimental Details}
\label{app:details}

\begin{table*}[t]
\centering
\small
\begin{tabular}{@{}lrrrrrrrr@{}}
\toprule
& & & & \multicolumn{4}{c}{stored share of offered bits (\%)} & bits/param \\
\cmidrule(lr){5-8}
Base model and adapter & params & deficit & $n$ & 2K & 4K & 8K & 16K & at 16K \\
\midrule
Pythia-70M, LoRA all, $r{=}1$ & 49K & 1254 & 70 & 0 & 2 & 35 & 73 & 0.73 \\
 &  &  & 140 & 0 & 1 & 15 & 29 & 0.58 \\
 &  &  & 280 & 0 & 0 & 0 & 0 & 0.00 \\
 &  &  & 560 & 0 & 0 & 0 & 0 & 0.00 \\
\addlinespace
Pythia-160M, LoRA all, $r{=}1$ & 147K & 1214 & 210 & 44 & 75 & 85 & 87 & 0.87 \\
 &  &  & 420 & 0 & 22 & 44 & 51 & 1.02 \\
 &  &  & 840 & 2 & 12 & 20 & 26 & 1.05 \\
 &  &  & 1{,}680 & 0 & 0 & 0 & 0 & 0.00 \\
\addlinespace
Pythia-410M, LoRA all, $r{=}1$ & 393K & 331 & 560 & 81 & 88 & 91 & 94 & 0.94 \\
 &  &  & 1{,}120 & 42 & 46 & 60 & 66 & 1.32 \\
 &  &  & 2{,}240 & 0 & 12 & 19 & 21 & 0.83 \\
 &  &  & 4{,}480 & 0 & 0 & 0 & 0 & 0.00 \\
\addlinespace
Qwen2.5-0.5B, LoRA all, $r{=}1$ & 550K & 364 & 800 & 98 & 98 & 98 & 98 & 1.01 \\
 &  &  & 1{,}600 & 77 & 86 & 91 & 93 & 1.91 \\
 &  &  & 3{,}200 & 28 & 40 & 48 & 54 & 2.22 \\
 &  &  & 6{,}400 & 0 & 0 & 0 & 0 & 0.00 \\
\addlinespace
Qwen2.5-1.5B, LoRA all, $r{=}1$ & 1154K & 353 & 1{,}600 & 98 & 98 & 98 & 98 & 0.96 \\
 &  &  & 3{,}200 & 71 & 85 & 91 & 94 & 1.84 \\
 &  &  & 6{,}400 & 13 & 31 & 44 & 51 & 2.00 \\
\addlinespace
Qwen2.5-0.5B, LoRA MLP, $r{=}1$ & 415K & 364 & 1{,}200 & 82 & 89 & 92 & 94 & 1.92 \\
 &  &  & 2{,}400 & 34 & 44 & 51 & 56 & 2.29 \\
\addlinespace
Qwen2.5-0.5B, LoRA attn., $r{=}3$ & 406K & 364 & 1{,}200 & 7 & 21 & 0 & 0 & 0.00 \\
 &  &  & 2{,}400 & 0 & 0 & 0 & 0 & 0.00 \\
\bottomrule
\end{tabular}
\caption{Rank-1 LoRA and matched placements on pretrained language models (batches of 64 sequences, learning rate $3\times10^{-3}$).}
\label{tab:lm}
\end{table*}

\begin{table}[t]
\centering
\small
\begin{tabular}{@{}lrr@{}}
\toprule
Adapter on the 2.2M base & rank & parameters \\
\midrule
Attention (fused QKV, output) & 8 & 36{,}864 \\
All four projections & 3 & 36{,}864 \\
MLP (both projections) & 5 & 38{,}400 \\
All four projections & 1 & 12{,}288 \\
All four projections & 16 & 196{,}608 \\
Prompt tuning (190 vectors) & -- & 36{,}480 \\
Full fine-tuning & -- & 2{,}215{,}488 \\
\bottomrule
\end{tabular}
\caption{Trainable parameters of the configurations on the 2.2M base. Rank-1 and rank-16 LoRA on the 8.1M base have 30{,}720 and 491{,}520 parameters.}
\label{tab:params}
\end{table}

\paragraph{Small GPT-style bases.}
The two small bases are pre-norm decoders with learned positional embeddings, GELU MLPs of width $4d$, and output weights tied to the input embeddings. The 2.2M-parameter base has 4 layers, 4 heads and width $d=192$; the 8.1M-parameter base has 6 layers, 5 heads and width 320. Both use a byte-level BPE vocabulary of 2{,}048 tokens trained on WikiText-103 plus one start token, and were trained for 200M tokens (about 1.7 passes over a 120M-token sample of WikiText-103) with AdamW, peak learning rate $10^{-3}$, 200 warmup steps, and batches of 256 sequences of 256 tokens. Their final validation losses are 3.28 and 2.87 nats per token. For the comparison of frozen bases in Section~\ref{sec:factors} we also use a randomly initialized copy of the 2.2M architecture (weights drawn from $\mathcal{N}(0,0.02^2)$ with a fixed seed) and a copy pretrained for 30M tokens on a synthetic first-order Markov chain over the same 2{,}048 symbols, with Zipf-distributed marginals and 24 successors per symbol.

\paragraph{Random sequences.}
Each training sequence is a start token followed by 64 tokens drawn uniformly from the 2{,}048 payload symbols; only the 64 payload tokens are scored. Held-out sequences are drawn from the same distribution with a different seed. On Qwen2.5 we draw payload tokens from ids 1{,}000 to 3{,}047 and use the end-of-text token as the start token.

\paragraph{Adapters.}
LoRA uses scale $a=2r$ and no dropout. $B$ starts at zero; on the small bases $A$ is drawn from $\mathcal{N}(0,1/d_{\mathrm{in}})$, and on Qwen2.5 we keep the PEFT default. On the small bases, attention placement adapts the fused query--key--value projection and the output projection of every layer, MLP placement adapts both MLP projections, and all placement adapts all four. Table~\ref{tab:params} lists the resulting parameter counts. Prompt tuning prepends 190 trained vectors to the input (36{,}480 parameters on the 2.2M base), initialized from the embeddings of randomly chosen tokens \citep{lester2021power}. On Qwen2.5, LoRA adapts the query, key, value, output, gate, up and down projections of every layer.

\paragraph{Training.}
All capacity runs use AdamW with $\beta=(0.9,0.95)$, no weight decay, and gradient clipping at 1.0. Peak learning rates are $3\times10^{-3}$ for LoRA and prompt tuning on the small bases ($10^{-3}$ for rank 16 on the 8.1M base, where $3\times10^{-3}$ diverged in earlier runs), $10^{-3}$ for full fine-tuning, and $3\times10^{-3}$ on the pretrained language models ($10^{-3}$ for the placement comparison on Qwen2.5-0.5B). The schedule has 500 warmup steps (200 on Qwen2.5), then a constant rate; each budget $\steps$ is read from a copy of the run whose rate decays linearly to zero between steps $0.8\steps$ and $\steps$. Adapter parameters are kept in fp32. Small-base runs use TF32 matrix multiplication during training and full fp32 for scoring; the Qwen2.5 base weights are kept in bf16.

\paragraph{SFT and GRPO.}
Both arms train a rank-4 LoRA ($a=8$, no dropout) on all linear layers of Qwen2.5-0.5B-Instruct for three passes over the training prompts. SFT uses AdamW with learning rate $10^{-4}$, batches of 8, and a loss on the answer tokens followed by the end-of-text token. GRPO uses TRL 0.15.2 with learning rate $10^{-5}$, 4 prompts per step, 8 samples per prompt at temperature 0.9, at most 64 generated tokens, and KL coefficient 0.04. An answer earns reward 1 if the first line of the completion starts with the target string. Accuracy is the share of the first 200 held-out prompts whose greedy completion starts with the target.

\paragraph{Learning rates and seeds.}
For LoRA on the small bases we use the better of $10^{-3}$ and $3\times10^{-3}$ from earlier 16K-step runs; for the pretrained language models we compared $3\times10^{-4}$, $10^{-3}$ and $3\times10^{-3}$ on Qwen2.5-0.5B with 1{,}600 sequences and 4K steps (57\%, 76\% and 86\% of the bits stored) and used $3\times10^{-3}$. The six configurations of Figure~\ref{fig:capacity}a are run with three seeds, each seed drawing its own random sequences; tables report the mean and the sample standard deviation. The SFT, GRPO and control adapters are also run with three seeds. The frozen-base comparison, the logit-scale runs, the 8.1M base and the pretrained language models are single runs.

\paragraph{Artifacts and licenses.}
WikiText-103 \citep{merity2017pointer} is distributed under CC BY-SA 3.0 and GFDL; we use it only to pretrain the two small bases. Qwen2.5 \citep{qwen2025qwen25} and Pythia \citep{biderman2023pythia} models are released under the Apache 2.0 license, as are Transformers, PEFT and TRL; PyTorch uses a BSD license. The icons in Figure~\ref{fig:overview} are from Microsoft's Fluent Emoji set (MIT license). All uses are research uses consistent with these licenses. All task data, including the personnel records and secrets, are generated by our scripts and name only fictitious users (``user00914'').

\paragraph{Software and compute.}
Experiments use PyTorch 2.1.2, Transformers 4.49.0, PEFT 0.14.0 and TRL 0.15.2 on NVIDIA A100, H100 and H200 GPUs; the runs took about 110 GPU-hours in total.

\section{Pretrained Language Models}
\label{app:pythia}

Table~\ref{tab:lm} lists the runs on the pretrained language models with learning rate $3\times10^{-3}$ and up to 16K steps: the share of the offered bits that the adapter stores after each budget, and the bits per parameter after 16K steps. The deficit column gives the mean codelength of a random sequence under the base model minus 704 bits.

For the Pythia models and Qwen2.5-0.5B, the largest dataset gives a reading of zero, and so do the two larger datasets of Pythia-70M. On Qwen2.5-1.5B, the largest dataset (6{,}400 sequences) gives the highest reading, so the 2.00 bits per parameter reported for it may be below its capacity. In these runs training still lowers the codelength of unseen random sequences, from 1{,}068 to about 840 bits on Qwen2.5-0.5B, so the adapter learns the overall statistics of random tokens. It does not bring any training sequence below 704 bits within 16K steps, and Equation~\ref{eq:mem} therefore counts nothing. The capacity estimates in Section~\ref{sec:capacity} use the dataset sizes with the largest readings; they are lower bounds, both because the budget is short and because the clipped measure ignores sequences that are partly but not fully learned.

\end{document}